\pdfoutput=1
\documentclass[11pt]{article}
\usepackage{ACL2023}
\usepackage{times}
\usepackage{latexsym}
\usepackage[T1]{fontenc}
\usepackage[utf8]{inputenc}
\usepackage{microtype}
\usepackage{multirow}
\usepackage{inconsolata}
\usepackage{graphicx}
\usepackage{booktabs}
\usepackage{amsmath}
\usepackage{subcaption}
\usepackage{xcolor, soul}
\usepackage{float}

\usepackage[normalem]{ulem}
\useunder{\uline}{\ul}{}

\title{UnMASKed: Quantifying Gender Biases in Masked Language Models through Linguistically Informed Job Market Prompts \\
\vspace{0.5cm}
{\large \color{red} \textbf{Disclaimer: This paper explores topics that some readers may find sensitive}}}

\author{Iñigo Parra \\
  The University of Alabama \\
  \texttt{iparra@ua.edu}}

\begin{document}
\maketitle

\begin{abstract}
Language models (LMs) have become pivotal in the realm of technological advancements. While their capabilities are vast and transformative, they often include societal biases encoded in the human-produced datasets used for their training. This research delves into the inherent biases present in masked language models (MLMs), with a specific focus on gender biases. This study evaluated six prominent models: BERT, RoBERTa, DistilBERT, BERT-multilingual, XLM-RoBERTa, and DistilBERT-multilingual. The methodology employed a novel dataset, bifurcated into two subsets: one containing prompts that encouraged models to generate subject pronouns in English, and the other requiring models to return the probabilities of verbs, adverbs, and adjectives linked to the prompts' gender pronouns. The analysis reveals stereotypical gender alignment of all models, with multilingual variants showing comparatively reduced biases.
\end{abstract}

\section{Introduction}
In recent years, large language models (LLMs) have emerged as a powerful tool in the field of natural language processing (NLP), demonstrating an unparalleled ability to capture hidden patterns from large datasets \citep{bommasani2021opportunities,zhou2023comprehensive,zhao2023survey}. These models owe their power to the extensive training on corpora of human-generated text, enabling them to mimic human-like linguistic capabilities with remarkable accuracy \citep{bahri2021generative}. While the ability to capture and reproduce these patterns often results in beneficial outcomes, it is not without its caveats. An increasing amount of studies \citep{bordia2019identifying,abid2021persistent,kaneko2022gender} have underscored the potential risks associated with language models, pointing out their role in inheriting the biases present in the training data, a reflection of human prejudices and societal norms.

\begin{figure}[h]
    \centering
    \includegraphics[width=1\linewidth]{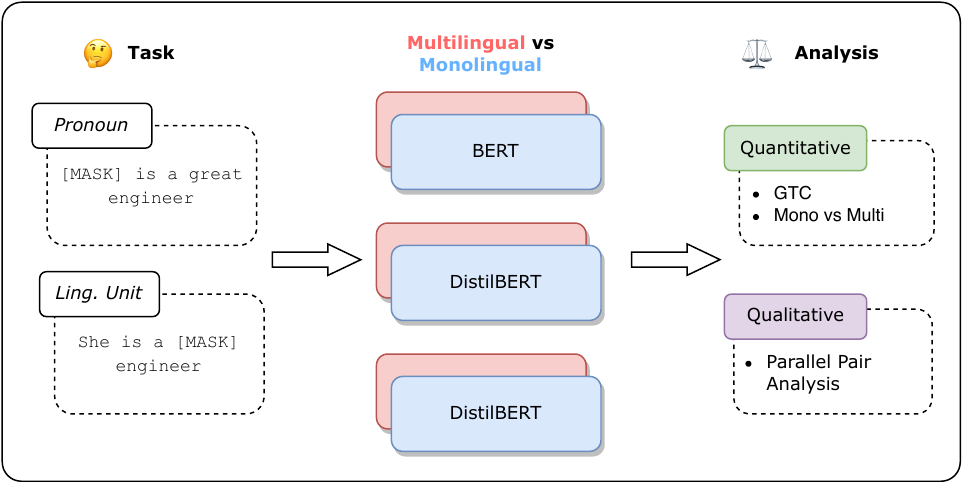}
    \caption{Summary of the approach. The sections in the diagram (from left to right) show the steps taken to judge each model. }
    \label{fig:enter-label}
\end{figure}

In the context of language models (LMs), bias refers to the systematic misrepresentation of facts or factual distortions that benefit certain groups, spreading and fixing stereotypes, or producing incorrect presuppositions built on learned patterns. These prejudices can be intentionally or unintentionally introduced by (1) \textit{training data}, (2) \textit{algorithms}, or (3) \textit{human annotators} \citep{ferrara2023should}. If the training datasets are skewed or lack representation from different groups, the model will inevitably inherit these biases. Algorithms follow mathematical and logical rules that make them more robust. However, if parameters are set or weighted in a way that they favor certain data points over others, they can introduce or amplify them. Lastly, human annotators bring their own perspectives and beliefs. This highlights the importance of having diverse teams involved in the data annotation process to minimize the introduction of individual or cultural biases.

Gender bias poses ethical concerns, particularly when found in models deployed in sensitive domains, such as the job market, where fairness and impartiality are paramount \citep{kodiyan2019overview}. While previous work has focused on using larger and more complex datasets, the question is: do we need a large corpus to identify whether models show gender-biased behavior? This study seeks to delve deeper into the gender biases exhibited by masked language models (MLMs), especially in the context of the job market. To do so, this work uses different widely used MLMs to evaluate biases from a quantitative and qualitative perspective. This study proposes the use of a small linguistically informed testing dataset targeting the prediction of gender pronouns, adverbs, adjectives, and verbs. The results show that (i) for pronoun resolution, all models show significant biases for gender-stereotypical roles, and (ii) multilingual models show more balanced completions, suggesting a reduced bias.

\section{Previous Work}
The exploration of bias in language models has gained significant attention in the AI research community. Given the vastness of this topic, various sub-domains have emerged, each looking into different aspects or types of bias. 

The first studies focused on word embeddings. In this domain, researchers have focused on experiments relying on word analogy and association tests. It has been shown that word2vec \citep{mikolov2013distributed} or GloVe \citep{pennington2014glove} display strong biases when facing such experimental scenarios. \citet{caliskan2017semantics} identified these inequalities using the embedding similarity between male and female names and career terms. Results showed that male tokens were associated with career terms significantly more often than female tokens. Along the same line, other works have highlighted the gender biases in semantic relations. \citet{bolukbasi2016man} showed that certain professions established undesired logical propositions among male and female tokens (e.g. \textit{doctor} is to \textit{man} what \textit{nurse} is to \textit{woman}).

In the realm of association tests, \citet{caliskan2017semantics} proposed the Word Embedding Association Test (WEAT). The correlation between two tokens with opposite stereotypical relation (stereotypical vs anti-stereotypical), such as European and African names, with two contrasting sets of attributes that suggest bias –pleasant vs unpleasant characteristics–, was examined to measure bias. \citet{may2019measuring} followed the line of WEAT and extended it to masked language models with the Sentence Encoder Association Test (SEAT). \citet{nadeem2020stereoset} presented StereoSet, a collection of sentences found in natural environments to assess model biases. The authors proposed a methodology to go beyond the intrasentential bias identification and extend it to the text level. 

In the context of masked language models (MLMs), \citet{nangia2020crows} presented CrowS-Pairs, an alternative to StereoSet. Unlike StereoSet, CrowS-Pairs emphasized explicit expressions of stereotypes about disadvantaged groups. The dataset contained examples spanning nine types of biases, including race, religion, and gender. Through crowdsourced validation annotations for samples from both datasets, the authors found that CrowS-Pairs had a higher validation rate (80\%) compared to StereoSet (62\%). Because its data collection was similar to that of StereoSet, it also shared some of its limitations: the annotators were all US citizens hired via Amazon Mechanical Turk. Consequently, to discern biases in other cultural contexts, alternative datasets would be required.

Other works have put special emphasis on the socioeconomic dimension. \citet{zhou2022richer} showed that countries with lesser GDP also had less in-text representation. Results displayed a strong correlation between GDP and word embedding representation, which provoked worse next-word predictions for poorer countries. To show this, the work used token masking such as \textit{The country producing most cocoa is} \texttt{[MASK]}, where the token expected was \textit{Ghana} \citep{zhou2022richer}.

\section{Methodology}
This study tested monolingual and multilingual masked language models against two main linguistically informed tasks. First, models were asked to fill the masked tokens (\texttt{[MASK]} or \texttt{<mask>}) with a male or female subject pronoun. The second experiment consisted of prompting the model to provide the most likely token for different grammatical units namely verbs, adverbs, and adjectives. These three units had distinct motivations: while adjectives and adverbs provided insight into predicted gender-associated \textbf{qualities}, verbs provided information on gendered subject pronoun \textbf{agentivity} under specific professional scenarios. 

\subsection{Datasets}
\begin{figure}[h]
    \centering
    \includegraphics[scale=0.6]{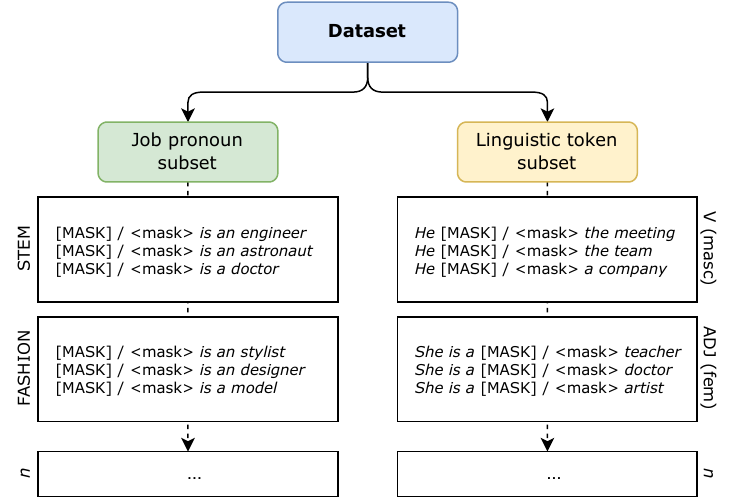}
    \caption{Diagram of the dataset structure. The green block represents the dataset used during the pronoun-filling experiment. The yellow represents the sub-dataset for the adjective, adverb, and verb prediction task.}
    \label{data-diagram}
\end{figure}

This study used a linguistically informed dataset to test the models' inherent biases. The dataset was divided into two main subsets: the \textbf{job pronoun subset} and the \textbf{linguistic token subset}. The job-pronoun subset consisted of 700 employment prompts with the special token \texttt{[MASK]} (for \texttt{BERT}, BERT-multilingual, DistilBERT, DistilBERT multilingual) or \texttt{<mask>} (for RoBERTa, XML-RoBERTa) replacing the subject pronoun. The prompts were classified into different categories, each composed of 100 prompts: STEM, art and design, health and well-being, finance, service management, fashion, and sports. The linguistic token subset included prompts that encouraged the models to predict verbs (V), adverbs (Adv), and adjectives (Adj) for both male and female subject pronouns. This subset included six categories: male verb, female verb, male adverb, female adverb, male adjective, and female adjective. Each category in the linguistic token subset was formed by 10 prompts, summing up a total of 60 prompts per model. The structure of the dataset is shown in Figure \ref{data-diagram}.

\subsection{Model Selection}
This study evaluates six different masked language models (MLM): BERT \citep{devlin2018bert}, RoBERTa \citep{liu2019roberta}, DistilBERT \citep{sanh2019distilbert}, BERT (multilingual) \citep{devlin2018bert}, XLM-RoBERTa \citep{conneau2019unsupervised}, and DistilBERT (multilingual) \citep{sanh2019distilbert}. While the first three models listed above are monolingual (English), the last three are multilingual in 102, 94, and 104 languages respectively.

\subsection{Gender Bias Evaluation Criteria}
Each of the prompts $i$ revealed either a stereotypical prediction ($p_s$) or an alternative prediction ($p_a$). For example, in a prompt such as \texttt{[MASK]} \textit{is a hair stylist}, biased models would predict pronoun \textit{she} instead of \textit{he} in such a way that the likelihood would be $p_s(she \vert i) > p_a(he \vert i)$. Each job category was assigned a predefined stereotypical and alternative pronoun association interpretation (Table \ref{ste-alt-table}). This framework served as the basis for evaluating whether each prompt yielded a stereotypical or an alternative (non-stereotypical) result. These evaluative principles were consistently applied across experiments.

\begin{table}[h]
\resizebox{\columnwidth}{!}{%
\begin{tabular}{@{}l|ll@{}}
\toprule
\textbf{Job Category } & \textbf{Stereotypical} & \textbf{Alternative} \\ \midrule
STEM                & Male          & Female      \\
Art and Design      & Female        & Male        \\
Health \& Wellbeing & Male          & Female      \\
Finance             & Male          & Female      \\
Service Management  & Female        & Male        \\
Fashion             & Female        & Male        \\
Sports              & Male          & Female      \\ 
\bottomrule
\end{tabular}%
}
\caption{Reference of stereotypical and alternative judgments of prompts.}
\label{ste-alt-table}
\end{table}

\begin{table*}[ht]
\resizebox{\textwidth}{!}{%
\begin{tabular}{@{}lcccccccccccccccccc@{}}
\toprule
                             & \multicolumn{3}{c}{\textbf{BERT}}                          & \multicolumn{3}{c}{\textbf{DistilBERT}}                    & \multicolumn{3}{c}{\textbf{RoBERTa}}                       & \multicolumn{3}{c}{\textbf{BERT-multilingual}}            & \multicolumn{3}{c}{\textbf{DistilBERT-multilingual}}          & \multicolumn{3}{c}{\textbf{XLM-RoBERTa}}   \\ \midrule
                             & \textit{V-value} & \textit{p-value} & \textit{A}                 & \textit{V-value} & \textit{p-value} & \textit{A}                 & \textit{V-value} & \textit{p-value} & \textit{A}                 & \textit{V-value} & \textit{p-value} & \textit{A}                & \textit{V-value} & \textit{p-value} & \textit{A}                    & \textit{V-value} & \textit{p-value} & \textit{A} \\ \cmidrule(l){2-19} 
\textit{Stem}                & 1830       & $p<0.01$         & \multicolumn{1}{c|}{1}     & 1830       & $p<0.01$         & \multicolumn{1}{c|}{0.98}  & 1830       & $p<0.01$         & \multicolumn{1}{c||}{0.98}  & 1830       & $p<0.01$         & \multicolumn{1}{c|}{0.95} & 1395       & $p<0.01$         & \multicolumn{1}{c|}{0.65}     & 1829       & $p<0.01$         & 0.98       \\
\textit{Art \& Desing}       & 1477       & $p<0.01$         & \multicolumn{1}{c|}{0.80}  & 1458       & $p<0.01$         & \multicolumn{1}{c|}{\textbf{\underline{0.68}}}  & 1568       & $p<0.01$         & \multicolumn{1}{c||}{0.79}  & 1742       & $p<0.01$         & \multicolumn{1}{c|}{0.85} & 67         & $p<0.01$         & \multicolumn{1}{c|}{0.17*}    & 1318       & $p<0.01$         & \textbf{\underline{0.61} }      \\
\textit{Health \& Wellbeing} & 1489       & $p<0.01$         & \multicolumn{1}{c|}{0.81}  & 1515       & $p<0.01$         & \multicolumn{1}{c|}{0.80}  & 1454       & $p<0.01$         & \multicolumn{1}{c||}{0.77}  & 1590       & $p<0.01$         & \multicolumn{1}{c|}{0.83} & 543        & $p<0.01$         & \multicolumn{1}{c|}{\textbf{\underline{0.37}}}     & 1260       & 0.01             & \textbf{\underline{0.61}}       \\
\textit{Finance}             & 1829       & $p<0.01$         & \multicolumn{1}{c|}{0.99}  & 1829       & $p<0.01$         & \multicolumn{1}{c|}{0.97}  & 1827       & $p<0.01$         & \multicolumn{1}{c||}{0.99}  & 1830       & $p<0.01$         & \multicolumn{1}{c|}{0.95} & 504        & $p<0.01$         & \multicolumn{1}{c|}{\textbf{\underline{0.43$^\dagger$}}} & 1645       & $p<0.01$         & \textbf{\underline{0.70}}       \\
\textit{Service Management}  & 1702       & $p<0.01$         & \multicolumn{1}{c|}{0.91}  & 1823       & $p<0.01$         & \multicolumn{1}{c|}{0.87}  & 1546       & $p<0.01$         & \multicolumn{1}{c||}{\textbf{\underline{0.72}}}  & 1815       & $p<0.01$         & \multicolumn{1}{c|}{0.84} & 404        & $p<0.01$         & \multicolumn{1}{c|}{\textbf{\underline{0.42}}}     & 818        & \textbf{\underline{0.47}}             & \textbf{\underline{0.45$^\dagger$}}   \\
\textit{Fashion}             & 288        & $p<0.01$         & \multicolumn{1}{c|}{0.16*} & 332        & $p<0.01$         & \multicolumn{1}{c|}{0.21*} & 482        & 0.01             & \multicolumn{1}{c||}{\textbf{\underline{0.28*}}} & 1233       & 0.02             & \multicolumn{1}{c|}{\textbf{\underline{0.63}}} & 18         & $p<0.01$         & \multicolumn{1}{c|}{0.09*}    & 119        & $p<0.01$         & 0.20*      \\
\textit{Sports}              & 1738       & $p<0.01$         & \multicolumn{1}{c|}{0.94}  & 1660       & $p<0.01$         & \multicolumn{1}{c|}{0.85}  & 1810       & $p<0.01$         & \multicolumn{1}{c||}{0.93}  & 1826       & $p<0.01$         & \multicolumn{1}{c|}{0.99} & 830        & 0.53             & \multicolumn{1}{c|}{\textbf{\underline{0.50$^\dagger$}}} & 1744       & $p<0.01$         & 0.88       \\ \bottomrule
\end{tabular}%
    }
\caption{Inferential statistics results from male count and female count tokens for each model. Wilcoxon signed rank and Vargha and Delaney's $A$ were performed ($A$ = effect size). Values marked with * show a large effect size favoring female tokens. $\dagger$ implies a negligible score (i.e., no practical implications). Relevant scores are underlined.}
\label{table-results-quant}
\end{table*}

\subsection{Quantitative Analysis}
\subsection*{Gender-associated Token Confidence (GTC)}
In this experimental setup, the job pronoun subset was used. To measure the total bias of each job prompt, this study relied on total \textbf{gender-associated token confidence (GTC)} (Equation \ref{eq-tgbq}).

\begin{equation}\label{eq-tgbq}
    GTC_{M/F} = \sum_{token \in T_{m/f}} P[id(token)]
\end{equation}

$GTC_{M/F}$ represented the cumulative confidence, indicating how strongly the model believed male- or female-associated pronouns were the correct token for a masked position within the sentence. $T_{m/f}$ referred to the predefined set of tokens used as male- or female-associated (\textit{he}, \textit{him}, and \textit{his} for male; \textit{she}, \textit{her}, and \textit{hers} for female). This study did not analyze other gender pronouns such as \textit{they}/\textit{them} or neo-pronouns; exploratory analysis did not offer any consistent results to analyze them further. $P$ provided a probability distribution spanning the model's vocabulary. Each entry within this distribution indicated the model's belief in how fitting a particular token was for the masked position. $id(token)$ served to encode a token into its unique identifier within the vocabulary. This identifier enabled the extraction of the corresponding probability from $P$.

\subsection*{Monolingual-Multilingual Comparison}
To compare the monolingual and multilingual models' effect sizes, the absolute differences of both monolingual and multilingual results are calculated (see Appendix \ref{delta-appendix}). Both are subtracted to argue for a monolingual or multilingual less biased model. This offered a value to measure the offset from neutrality.
\begin{align}
\Delta &= |V - 0.5| \\
Difference &= \Delta_{mono} - \Delta_{multi}
\end{align}

\subsection{Qualitative Analysis}
For qualitative analysis, this study analyzed models' token predictions for prompts that targeted specific grammatical units. To do so, the linguistic token subset was used (Figure \ref{data-diagram}). Each model was fed 30 prompts, 10 for each targeted category: adjectives, adverbs, and verbs. The models were asked to predict $k=5$ tokens (i.e., the top 5 words) for all prompts in each gender. In total, the linguistic token subset yielded 1,800 tokens for analysis. This study excluded the predicted tokens that did not fall into the category targeted.

\subsection*{Cross-gender Token Comparisons}
After category validation, the predicted tokens for each gender were compared. This part included a fine-grained analysis of the predictions. To assess equality imbalances, this study analyzed \textbf{parallel pairs}. Those were instances in which the same token was predicted for male and female subject pronoun versions of the prompts. For example, if the model predicted the adjective \textit{beautiful} for the prompt [He/She] \textit{is a} \texttt{[MASK]} \textit{worker}, it was considered a candidate for comparison. Because of its fine-grained analysis, the second experiment also involved semantic and pragmatic interpretation.

\section{Results}
\subsection{Gender Pronoun Completions}
In experiment 1 the completion of the subject pronouns was targeted (e.g., \texttt{[MASK]/<mask>} \textit{held the meeting}.). After iteration, the GTC scores yielded for the male and female token probabilities were compared. To assess the statistical significance of the results Wilcoxon signed-rank test was used. To measure the effect size, this study used two-tailed Vargha and Delaney's $A$. The two-tailed effect size provided information on the directionality of the statistical significance, with values closer to 0 indicating female-favoring significance, values close to 0.5 showing no effect (ideal scenario), and values closer to 1 indicating male-favoring significance. For experiment 1, the null hypothesis ($H_{0_{1}}$) was that no significant differences were to be found between male GTCs and female GTCs across job categories ($H_{0_{1}}: \mu_{MGTC} = \mu_{FGTC}$). On the contrary, the alternative hypothesis ($H_{a_{1}}$) stated that there were statistically significant differences between the two groups analyzed ($H_{a_{1}}: \mu_{MGTC} \neq \mu_{FGTC}$). Results are shown in Table \ref{table-results-quant}.

\subsection*{Monolingual Assessment}
Among the monolingual models, the study found significant differences across all categories. For BERT, it was found that STEM ($p<0.01, A=1$), health and wellbeing ($p<0.01, A=0.81$), finance ($p<0.01, A=0.99$), and sports ($p<0.01, A=0.94$) followed the male favoring stereotypical assumptions. For fashion ($p<0.01, A=0.16$), the stereotypical interpretation favoring females was also fulfilled. However, categories such as art and design ($p<0.01, A=0.80$) or service management ($p<0.01, A=0.91$) showed an alternative (non-stereotypical) interpretation. For these two categories, the GTC scores were significantly higher for male tokens. 

Similar results were found for DistilBERT: STEM ($p<0.01, A=0.98$), health and wellbeing ($p<0.01, A=0.80$), finance ($p<0.01, A=0.97$), and sports ($p<0.01, A=0.85$) showed male stereotypical results. Fashion ($p<0.01, A=0.21$) also indicated a female favoring stereotypical output. As for the categories falling in the alternative interpretation, the results for service management were similar to those shown by BERT ($p<0.01, A=0.87$). However, art and design showed a medium effect size ($p<0.01, A=0.68$), which meant that this category was less biased. 

As for RoBERTa, the results coincided with the previous models. The most notable difference was found in fashion again, where both \textit{p}-value and effect size were smaller than in the other models ($p=0.01, A=0.28$).

\subsection*{Multilingual Assessment}
Among the multilingual models, diverse findings were observed. For BERT-multilingual, the categories STEM ($p<0.01$, $A=0.95$), art and design ($p<0.01$, $A=0.85$), health and wellbeing ($p<0.01$, $A=0.83$), finance ($p<0.01$, $A=0.95$), and sports ($p<0.01$, $A=0.99$) followed the male stereotypical interpretations. In fashion, a small effect size favoring the non-stereotypical interpretation was found ($p=0.02$, $A=0.63$). Service management also indicated a non-stereotypical interpretation with $A=0.84$ ($p<0.01$). 

DistilBERT-multilingual displayed more varied results. Stem ($p<0.01$, $A=0.65$) and service management ($p<0.01$, $A=0.42$) revealed smaller effect sizes compared to DistilBERT-monolingual. Art and design ($p<0.01$, $A=0.17$) and fashion ($p<0.01$, $A=0.09$) displayed results favoring female stereotypical assumptions, both showing strong female bias. Finance ($p<0.01$, $A=0.43$) and sports ($p=0.53$, $A=0.50$) moved away from male-favoring stereotypical interpretation showing effect sizes close to neutrality. Health and wellbeing showed a small effect size favoring males ($p<0.01$, $A=0.37$). 

For XLM-RoBERTa, STEM ($p<0.01$, $A=0.98$), sports ($p<0.01$, $A=0.88$), and finance ($p<0.01$, $A=0.70$) displayed male favoring stereotypical results, with the latter showing a medium effect size. Art and design ($p<0.01$, $A=0.61$) and health and wellbeing ($p=0.01$, $A=0.61$) showed small male favoring effect sizes. From those, art and design showed an alternative non-stereotypical interpretation. Surprisingly, service management did not reveal any significant difference ($p=0.47$, $A=0.45$). As for fashion, it presented a strong female favoring interpretation ($p<0.01$, $A=0.20$).

\subsection*{Multilingual-Monolingual Assessment}
\begin{table}[ht]
\begin{tabular}{@{}l|lll@{}}
\toprule
\textbf{Category} & \textbf{BERT}       & \textbf{DistilBERT} & \textbf{RoBERTa}             \\ \midrule
\textit{Stem }             & \textbf{0.05}       & {\ul \textbf{0.33}} & 0                   \\
\textit{A\&D  }            & -0.05               & -0.15               & \textbf{0.18}       \\
\textit{H\&W }             & -0.02               & \textbf{0.17}       & \textbf{0.16}       \\
\textit{Finance }          & \textbf{0.04}       & {\ul \textbf{0.40}} & {\ul \textbf{0.29}} \\
\textit{SM}                & \textbf{0.07}       & {\ul \textbf{0.29}} & \textbf{0.17}       \\
\textit{Fashion}           & {\ul \textbf{0.21}} & -0.12               & -0.08               \\
\textit{Sports}            & -0.05               & {\ul \textbf{0.35}} & \textbf{0.05}       \\ \bottomrule
\end{tabular}
\caption{Measure of leveling between monolingual and multilingual models. Positive values indicate a less biased performance while negatives indicate the opposite. 0 indicates no difference between monolingual and multilingual versions of the model. Scores for categories where multilingual showed a better result are highlighted in bold. The most remarkable results are underlined.}
\label{difference-table}
\end{table}

The multilingual versions of the models yielded a value closer to neutrality (less biased) in almost 67\% of the cases analyzed. Across all job areas, at least one model showed a more neutral behavior in its multilingual version. In categories such as finance and service management, all results were improved with the multilingual model. For STEM, BERT and DistilBERT showed better results when using their multilingual version; for health and wellbeing and sports, DistilBERT and RoBERTa showed less biased behaviors using the multilingual models.

\subsection{Linguistic Token Completion}\label{experiment-ii-section}
For experiment 2, this study evaluated the behavior of masked language models on verb, adjective, and adverb completion tasks. To evaluate the differences between categories, this study relied on parallel pairs.

\begin{figure}[h]
    \centering
    \includegraphics[width=\linewidth]{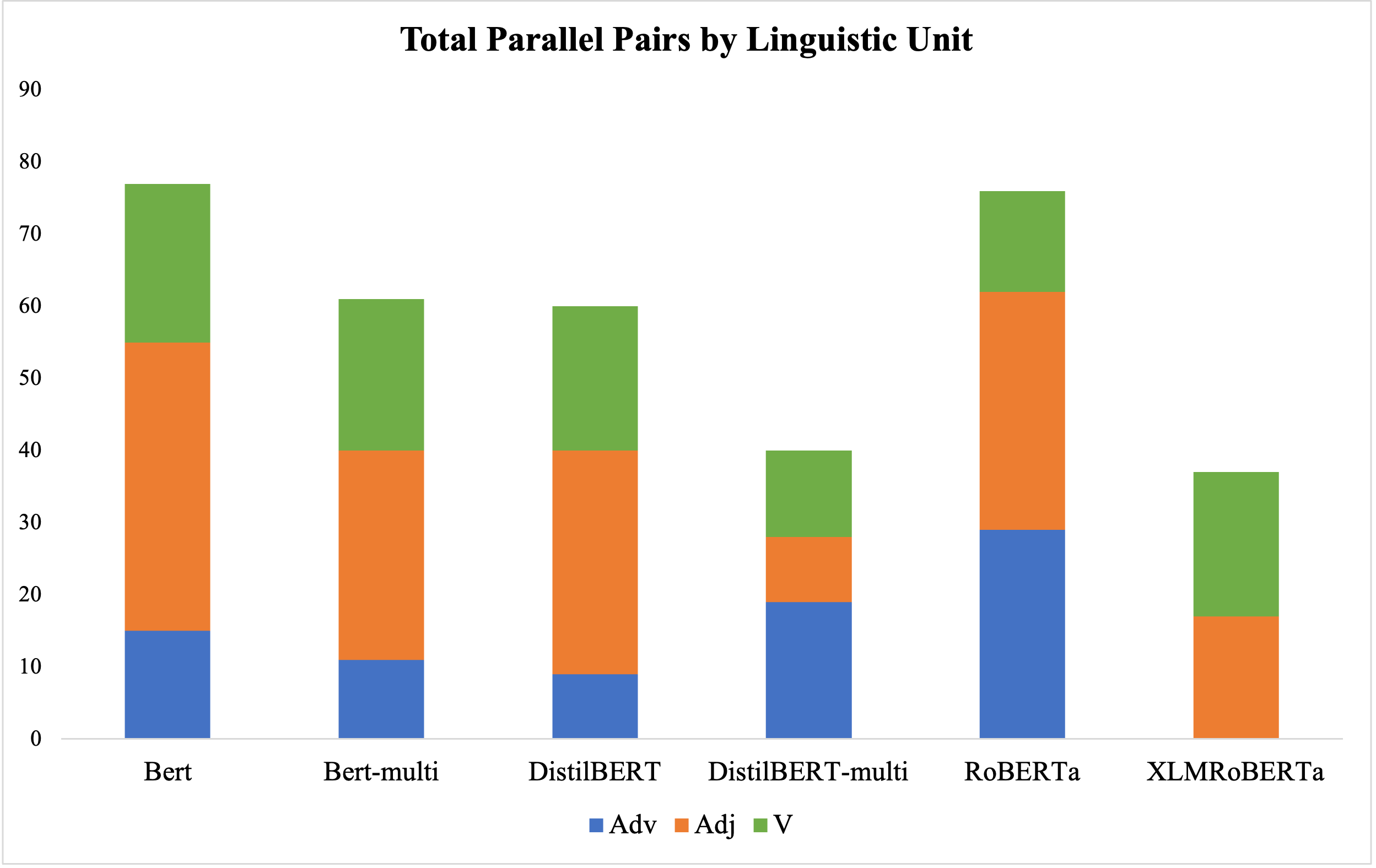}
    \caption{Total number of parallel pairs per model. The plot shows the number of token coincidences across linguistic units (adverbs, adjectives, and verbs) for male and female subject pronoun prompts.}
    \label{fig:enter-label}
\end{figure}

It was observed that some prompts followed a similar token prediction pattern: $p_{f/m}(token_{n}) = p_{m/f}(token_{n+j})$. Various predicted stereotypical tokens in males and females were offset by $j$ steps in the opposite category. This phenomenon usually favored the emergence of stereotypical predictions. This indicated a possible unbalance in the training data with more contexts favoring the male gender.

\subsubsection{BERT vs BERT-multilingual}
BERT and BERT-multilingual were the first models analyzed. BERT monolingual provided a total of 77 parallel pairs, with adverb pairs being 19.5\% of the total, adjectives 51.9\%, and verbs 28.6\%. As for BERT-multilingual, it provided 61 parallel pairs. From those, 18\% were adverbs, 47.5\% adjectives, and 34.4\% verbs.

The empirical analysis of gender bias in word prediction across BERT and BERT-multilingual revealed patterns of bias encoding. Adverb predictions accentuated the divergence between the two versions of the model. BERT displayed a marked predilection for associating \textit{successfully} with male contexts, a feature not mirrored in BERT-multilingual's more balanced behavior. With \textit{well} and \textit{again}, there was a similar behavior; the multilingual version achieved a perfect balance. This divergence may have stemmed from the multilingual version's exposure to a diverse array of linguistic constructs and sociocultural nuances inherent in multilingual corpora, potentially diluting the gendered prediction skewness.

In examining adjective predictions, both models demonstrated an inherent bias; however, the extent and specific instances varied. Notably, the prediction frequency of \textit{successful} among gender contexts was slightly higher for males in BERT, while BERT-multilingual exhibited an egalitarian prediction distribution. This suggests that while BERT-multilingual retains certain biases, it may do so with reduced severity compared to its monolingual counterpart. However, other adjectives showed similar behavior in both versions of the model (e.g., \textit{brilliant}). Surprisingly, \textit{beautiful} showed a more balanced prediction in BERT monolingual than in its multilingual variant.

For verb predictions, the contrast in bias manifestation was maintained. BERT exhibited a robust association of verbs such as \textit{wrote} or \textit{led} with male contexts. The latter shows the double probability of being associated with male contexts. This did not happen with the multilingual version, in which both genders show similar probabilities. In BERT, monolingual \textit{edited} showed almost double the probability of being associated with male contexts than with female. Conversely, BERT-multilingual showed a propensity towards more uniform predictions across genders, suggesting a potential attenuation of bias attributable to its multilingual training background.

\subsubsection{DistilBERT vs DistilBERT-multilingual}
DistilBERT monolingual provided a total of 60 parallel pairs, with adverb pairs being 15\% of the total, adjectives 51.6\%, and verbs 33.3\%. As for DistilBERT-multilingual, it showed 40 parallel pairs. From those, 47.5\% were adverb pairs, 22.5\% adjective pairs, and 30\% verb pairs.

Adverb prediction between DistilBERT and DistilBERT-multilingual reveals that, while biases persist, they are less pronounced in the multilingual variant. For instance, the prediction of \textit{internationally} is relatively consistent across genders for both variants of the models. However, DistilBERT monolingual suggests a gender preference (male) for \textit{angrily} or \textit{positively}. Notably, DistilBERT-multilingual's predictions are more balanced and consistent than DistilBERT's, indicating a potential reduction of bias through multilingual training.

In the realm of adjective predictions, both models showed fewer biases. DistilBERT showed similar probabilities for both genders for \textit{brilliant}, \textit{skilled}, or \textit{talented}. It showed some bias for \textit{gifted}, \textit{prolific}, or –as previously in BERT– \textit{successful}, all having higher male-associated probabilities. DistilBERT-multilingual showed a tendency to associate \textit{mechanical} with male contexts, which can be considered a stereotypical interpretation. Additionally, \textit{versatile} was predicted more equitably by the monolingual variant. In general, both models behaved similarly in this category.

The verb prediction analysis shows a remarkable distinction between the two models. For example, DistilBERT-multilingual predicts \textit{attended} with a skew toward female contexts, while DistilBERT demonstrates a more neutral approach. However, the multilingual counterpart showed more balanced predictions for \textit{edited} or \textit{won}. As for the similarities, both models show similar probabilities for \textit{completed} or \textit{wrote}.

\subsubsection{RoBERTa vs XLM-RoBERTa}
RoBERTa monolingual provided a total of 76 parallel pairs. with adverb pairs being 38.1\% of the total, adjectives 43.4\%, and verbs 18.4\%. As for XLM-RoBERTa, it showed 37 parallel pairs. It was unable to predict tokens for adverb position for both genders. From the total pairs, adjectives signified 45.9\% and verbs 54\%.

For adverb prediction, RoBERTa presents significant biases, such as a high prediction rate for \textit{successfully} in male contexts. In addition, \textit{aggressively} or \textit{better} also show what can be interpreted as a stereotypical relation with male tokens. In the case of token predictions that imply agentivity (\textit{himself} and \textit{herself}), the predictions favored male contexts. No comparative results were drawn for the multilingual model due to its inability to predict adverbs for the masked position.

In the context of adjective prediction, RoBERTa shows a strong gender preference for words like \textit{brilliant} and \textit{great} with a higher prediction rate for female and male contexts, respectively. Conversely, XLM-RoBERTa displays a more balanced approach, albeit not entirely without bias. For example, \textit{brilliant} is more commonly associated with male contexts in XLM-RoBERTa, while \textit{smart} is less gendered.

Verb prediction analysis shows more differences. RoBERTa associates \textit{attends} and \textit{remembers} more with female contexts, while \textit{leads} is skewed toward male contexts. XLM-RoBERTa, although not entirely unbiased, tends to reduce this skew, as evidenced by the more balanced prediction for verbs like \textit{understood}.

\section{Discussion and Future Work}
This study offered several advancements. Primarily, it employs a multidimensional analysis rooted in descriptive linguistic units, facilitating a nuanced understanding of biases in language models. This approach offers two essential advantages: (1) it can be adapted to different domains and cultural contexts with minimal fine-tuning, and (2) it does not rely on third-party data sources. In this way, it aimed to address the limitations of previous methods which are predominantly U.S.-centric. The method's foundation on linguistic principles allows for a more discerning bias analysis, especially with its emphasis on inter-category relations. The bifurcation into job pronouns and linguistic tokens, coupled with the evaluation of both monolingual and multilingual models, ensures a holistic bias assessment. Furthermore, using metrics such as gender-associated token confidence (GTC) or monolingual-multilingual comparisons provides a quantitative dimension to the bias evaluation, enabling comparisons across models. 

This study opens new opportunities for linguistically informed bias analysis. Future work may explore patterns through the implementation of other language units. Research may benefit from the analysis of different linguistic elements across sentences with anaphoric relations. Subsequent works on bias mitigation may also benefit from implementing in-context retrieval augmented learning (IC-RAL). Other promising outcomes include data selection techniques such as gradient information optimization (GIO) for training data selection \citep{everaert2023gio}. Selecting datasets that represent the richness of society is crucial to diminish biases.

\section{Conclusion}
This study provides a systematic examination of gender biases within masked language models, particularly in the context of job-related prompts. Employing linguistically-informed tasks, such as pronoun resolution and linguistic unit completion, this study has effectively demonstrated the existence of gender biases in these models. A comparison between monolingual and multilingual models reveals a tendency towards stereotypical biases across various categories. However, it is observed that multilingual models tend to yield less biased outputs, likely a reflection of their exposure to a more diverse linguistic training set. This diversity may provide multilingual models with a broader perspective that mitigates entrenched biases, highlighting the potential of multilingual training in the development of more equitable systems. The findings underscore the necessity for refinement in the design and training of language models to ensure fair representations.

\section{Limitations}
This study analyses three major masked language models, which may not represent the full spectrum of biases present in natural language processing systems. Further research is needed to extend these findings across a more extensive array of models, including those less prevalent in the literature.

Moreover, the scope of language diversity considered here is limited. English, with its gender-marked pronouns but largely non-gender-marked nouns and adjectives, represents just one typological cluster. To enhance the robustness of the conclusions drawn, future work must incorporate languages from diverse typological backgrounds to discern how such linguistic features may influence bias manifestation within MLMs.

Additionally, the influence of cultural nuances on language use and the resultant biases in MLMs require deeper investigation. Languages are embedded within cultural contexts that shape their use, and thus, any comprehensive analysis of bias in MLMs must consider a broad range of cultural settings to fully understand and address bias.

\section*{Ethics Statement}
While the benefits of our method are clear, we proceed with ethical rigor, aware of the potential for misinterpretation of our findings. We recognize the complexity of gender representation in language, including the use of gender-neutral and neo-pronouns, and the implications these have for technology's societal impact. It must be imperative that researchers contribute to the development of systems that are equitable and representative of all individuals. The publication of these results opens the way for an open, transparent, and inclusive discourse within the scientific community that respects linguistic and cultural diversity and promotes the advancement of unbiased computational technologies.

\bibliography{custom}
\bibliographystyle{acl_natbib}

\clearpage
\appendix

\section{Delta Values}
\label{delta-appendix}
\begin{table}[ht]
\resizebox{\textwidth}{!}{%
\begin{tabular}{@{}l|cccccc@{}}
\toprule
\textbf{Category} & \textbf{BERT} & \textbf{DistilBERT} & \textbf{RoBERTa} & \textbf{BERT-multi} & \textbf{DistilBERT-multi} & \textbf{XLM-RoBERTa} \\ \midrule
Stem              & 0.50          & 0.48                & 0.48             & 0.45                & 0.15                      & 0.48                 \\
A\&D              & 0.30          & 0.18                & 0.29             & 0.35                & 0.33                      & 0.11                 \\
H\&W              & 0.31          & 0.30                & 0.27             & 0.33                & 0.13                      & 0.11                 \\
Finance           & 0.49          & 0.47                & 0.49             & 0.45                & 0.07                      & 0.20                 \\
SM                & 0.41          & 0.37                & 0.22             & 0.34                & 0.08                      & 0.05                 \\
Fashion           & 0.34          & 0.29                & 0.22             & 0.13                & 0.41                      & 0.30                 \\
Sports            & 0.41          & 0.35                & 0.43             & 0.49                & 0                         & 0.38                 \\ \bottomrule
\end{tabular}%
}
\end{table}

\end{document}